\documentclass[oribibl]{llncs}
\usepackage{makeidx}  
\usepackage{times}
\usepackage{helvet}
\usepackage{courier}
\usepackage{amssymb}
\usepackage{amsfonts}
\usepackage{amsmath,bm}
\usepackage{enumerate}
\usepackage{mathtools}  
\usepackage{graphicx}
\usepackage{tikz}
\usetikzlibrary{shapes,arrows,automata}
\usepackage{latexsym}
\usepackage{subfigure}
\usepackage[ruled,linesnumbered]{algorithm2e}
\begin{document}
\newtheorem{excont}{Example}{\bfseries}{\itshape}
\renewcommand{\theexcont}{\theexample}
\newtheorem{axiom}{Axiom}{\bfseries}{\rmfamily}
\def\diag{\mathop{\rm diag}}

\title{Argument Ranking with Categoriser Function
\thanks{This work is supported by the Funds NSFC61171121, NSFC60973049, and the Science Foundation of Chinese Ministry of Education-China Mobile 2012.}}
%
%
\author{Fuan Pu \and Jian Luo \and Yulai Zhang \and Guiming Luo}
\authorrunning{Pu Fuan et al.} 
%
\tocauthor{Pu Fuan, Luo Jian, Zhang Yulai and Luo Guiming}
\institute{School of Software, Tsinghua University, Beijing, China,\\
\email{\{pfa12,j-luo10,zhangyl08\}@mails.tsinghua.edu.cn, gluo@mail.tsinghua.edu.cn}}

\maketitle              

\begin{abstract}
Recently, ranking-based semantics is proposed to rank-order arguments from the most acceptable to the weakest one(s), which provides a graded assessment to arguments. In general, the ranking on arguments is derived from the strength values of the arguments. Categoriser function is a common approach that assigns a strength value to a tree of arguments. When it encounters an argument system with cycles, then the categoriser strength is the solution of the non-linear equations. However, there is no detail about the existence and uniqueness of the solution, and how to find the solution (if exists). In this paper, we will cope with these issues via fixed point technique. In addition, we define the categoriser-based ranking semantics in light of categoriser strength, and investigate some general properties of it. Finally, the semantics is shown to satisfy some of the axioms that a ranking-based semantics should satisfy.

\end{abstract}

\section{Introduction}
The field of computational models of argumentation \cite{ref-rahwan2009argumentation} aims at reflecting on how human argumentation utilizes incomplete and inconsistent knowledge to construct and analyze arguments about the conflicting options and opinions.

The most popularly used framework to talk about general issues of argumentation is that of abstract argumentation \cite{ref-Dung1995AAF}, which provides a unifying and powerful tool for the study of many formal systems developed for common-sense reasoning. In the past nearly 20 years, several different kinds of semantics for abstract argumentation system have been proposed that highlight various aspects of argumentation \cite{ref-baroni2005scc,ref-dung2006dialectic,ref-baroni2010computational}. Those semantics partition the set of arguments into two classes: extensions and non-extensions. Each extension is a set of arguments, which is able to
``survive together'' and represents a coherent point of view. In order to reason with a semantics one has to take either a credulous or skeptical perspective. In other words, an argument is ultimately \textit{accepted} with respect to a semantics if it belongs to every extension; an argument is \textit{rejected} if it dose not belong to any extension; and an argument is \textit{undecided} if it is in some extensions and not in others.

However, those semantics may exhibit a variety of problematic aspects such as emptiness, non-existence, multiplicity \cite{ref-bench2007ArgAI} when encountering cycles, and are not suitable for practical applications in some scenarios. Considering an argument system whose grounded extension is empty, for example, if one must make a choice, then the grounded semantics is unavailable since all arguments are unacceptable in this case.

Recently, \cite{ref-amgoud2013ranking} introduces a new family of semantics, which provide a graded assessment to arguments, i.e., it ranks arguments from the most acceptable to the weakest one(s). In fact, this line of thinking has been mentioned in \cite{ref-cayrol2005graduality}, in which two approaches, generic local valuation and global valuation, are proposed to evaluate the strength of an argument, and then a preordering (ranking) on arguments is induced by those strength values. In particular, the authors show that the approach for local valuation generalizes the categoriser function \cite{ref-besnard2001logic}, and enables to handle cycles, then the strength valuation is the solution of second-degree equations. However, there is no detail about the following questions: Does there exist a solution for these equations? If it exists, is it unique or multiple and how to find them?

In this paper, we expect to tackle these issues by fixed-point technique. In addition, a ranking-based semantics, called categoriser-based ranking semantics, is defined in the light of categoriser valuation, and some of its properties are investigated. Lastly, we prove that the semantics satisfies some of the axioms, proposed by \cite{ref-amgoud2013ranking}, which a ranking-based semantics should satisfy. The remainder of this paper is structured as follows. In Section 2, we briefly recall some backgrounds on abstract argumentation and the ranking-based semantics for argumentation frameworks. In Section~3, we employ the fixed-point technique to analyze the categoriser strength valuation for argumentation system, and the categoriser-based ranking semantics is defined. We relate the semantic with \cite{ref-amgoud2013ranking} in Section~4 and conclude in Section~5.

\section{Preliminaries}

\subsection{Abstract Argumentation Framework}
Abstract argumentation frameworks \cite{ref-Dung1995AAF} convey a very simple view on argumentation since they do not presuppose any internal structure of an argument. Here, the interactions among arguments are attack relations, which express conflicts between them.

\begin{definition}[Abstract Argumentation Framework] \label{Def_Argumentation_Framework}
   An argumentation framework is a pair $\textit{AF}=\left< \mathcal{X}, \mathcal{R}\right>$ where $\mathcal{X}$ is a finite set of arguments and $\mathcal{R} \subseteq \mathcal{X} \times \mathcal{X}$ is a binary relation on $\mathcal{X}$, also called attack relation. $(a,b)\in \mathcal{R}$ means that $a$ attacks $b$, or $a$ is a (direct) attacker of $b$. Often, we write $(a,b) \in \mathcal{R}$ as $a \mathcal{R} b$.
\end{definition}

 We denote by $\mathcal{R}^-(x)$ (respectively, $\mathcal{R}^+(x)$) the subset of $\mathcal{X}$ containing those arguments that attack (respectively, are attacked by) the argument $x\in\mathcal{X}$, extending this notation in the natural way to sets of arguments, so that for $S\subseteq \mathcal{X}$, $\mathcal{R}^-(S) \triangleq \{x\in\mathcal{X}: \exists y \in S \mbox{~such that~} x\mathcal{R}y\}$ and $\mathcal{R}^+(S) \triangleq \{x\in\mathcal{X}: \exists y \in S \mbox{~such that~} y\mathcal{R}x\}$.

 A set $S\subseteq \mathcal{X}$ is \emph{conflict-free} iff $S \cap \mathcal{R}^-(S)=\emptyset$. Let $\mathfrak{F}: 2^\mathcal{X} \mapsto 2^\mathcal{X}$ be the \emph{characteristic function} of an argument system such that $\mathfrak{F}(S)=\{x\in\mathcal{X}: \mathcal{R}^-(x) \subseteq \mathcal{R}^+(S)\}$. We define the \emph{defenders} of an argument $x$, denoted by $\mathcal{D}(x)$, are the attackers of the elements of $\mathcal{R}^-
 (x)$. Formally, $\mathcal{D}(x)=\{y\in\mathcal{X}: y \in \mathcal{R}^-\big(\mathcal{R}^-(x)\big)\}$.

 To define the solutions of an argument system, we mean selecting a set of arguments that satisfy some acceptable criteria. Let $S\subseteq\mathcal{X}$ be conflict-free, then, $S$ is \textbf{admissible} iff $S\subseteq\mathfrak{F}(S)$; $S$ is a \textbf{preferred extension} iff it is a maximal (w.r.t. $\subseteq$) admissible set; $S$ is a \textbf{complete extension} iff $S=\mathfrak{F}(S)$; $S$ is a \textbf{grounded extension} iff it is the minimal (w.r.t. $\subseteq$) complete extension (or, alternatively, it is the least fixed point $\mathfrak{F})$; $S$ is a \textbf{stable extension} iff $\mathcal{R}^+(S)= \mathcal{X} \backslash S$.

\begin{example}\label{Exp_ArgNetwork}
Let us consider the abstract argumentation framework illustrated in Figure~\ref{Fig_ExampArg}, in which vertices represent arguments and direct arcs correspond to attacks (i.e. elements of $\mathcal{R}$). For this example, $\{x_1, x_3\}$ is the preferred, complete and grounded extension, however, there exist no stable extensions at all.
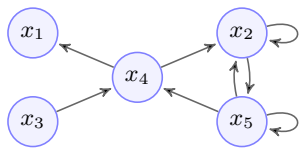
\begin{figure}[htb]
\vspace{-0.3cm}
\centering
  \begin{tikzpicture}[->,>=stealth',shorten >=1pt,auto,node distance=1.4cm, semithick]

\tikzstyle{vBlue}=[draw=blue!50,fill=blue!5,circle,text width=5.5mm,inner sep=1pt,minimum height=6pt, align=center]
\tikzstyle{every edge}=[draw=black!60]

\node[vBlue](x4){$x_4$};
\node[vBlue, above left of=x4, yshift=-4mm, xshift=-4mm](x1){$x_1$};
\node[vBlue, below left of=x4, yshift=4mm, xshift=-4mm](x3){$x_3$};
\node[vBlue, above right of=x4, yshift=-4mm, xshift=4mm](x2){$x_2$};
\node[vBlue, below right of=x4, yshift=4mm, xshift=4mm](x5){$x_5$};

\path (x4) edge                (x1)
      (x3) edge                (x4)
      (x4) edge                (x2)
      (x5) edge                (x4)
      (x5) edge [bend left=12] (x2)
      (x2) edge [bend left=12] (x5)
      (x5) edge [loop right]   (x5)
      (x2) edge [loop right]   (x2);
\end{tikzpicture}
\caption{A simple example of argumentation framework} \label{Fig_ExampArg}
\vspace{-0.6cm}
\end{figure}
\end{example}

From the above example, it is shown that an abstract argumentation framework can be represented as a digraph, known as attack graph. One of the often-used ways is to represent a digraph as 0-1 matrix for computational purposes. For an argumentation system $\textit{AF}=\left< \mathcal{X}, \mathcal{R}\right>$ with $\mathcal{X}=\{x_1,x_2,\cdots,x_n\}$, we define the \textbf{attack matrix} of $\textit{AF}$ as the $n\times n$ matrix $\mathbf{D}=[d_{ij}]$ such that $d_{ij}=1$ if $x_j\mathcal{R}x_i$; otherwise, $0$.\footnote{In fact, the attack matrix of an argumentation framework is the transpose of the adjacency matrix of its corresponding attack graph.} For instance, the attack matrix of the argument system in Example~\ref{Exp_ArgNetwork} is
\begin{equation*}
\mathbf{D}=
  \left[
    \begin{array}{ccccc}
      ~0~ & ~0~ & ~0~ & ~1~ & ~0~ \\
      0 & 1 & 0 & 1 & 1 \\
      0 & 0 & 0 & 0 & 0 \\
      0 & 0 & 1 & 0 & 1 \\
      0 & 1 & 0 & 0 & 1 \\
    \end{array}
  \right]
\end{equation*}
Moreover, we denote the $i$-th row of $\mathbf{D}$ by $\mathbf{D}_{i\star}$, which can indicate some information about the direct attackers of argument $x_i$, e.g., the sum of $\mathbf{D}_{i\ast}$ shows the number of attackers of $x_i$.

\subsection{Ranking-based semantics for argument system}
In order to provide a graded assessment to arguments, \cite{ref-amgoud2013ranking} proposes ranking-based semantics which rank-order the set of arguments from the most acceptable to the weakest one(s). This novel approach is distinct from the already existing semantics which assign an absolute status (\emph{accepted}, \emph{rejected} and \emph{undecided}) for each argument. It compares pairs of arguments in the light of their respective sets of attackers, and states which arguments is more acceptable than another.

Before proceeding let us first formally characterize what we mean by the statements ``ranking'' in light of linear orderings \cite{ref-altman2008axiomatic}.
\begin{definition}[Ranking]
  Let $\mathcal{T}$ be some set. A ranking $\succeq$ on $\mathcal{T}$ is a binary relation on $\mathcal{T}$ such that:
  \begin{itemize}
    \item $\succeq$ is total (i.e. for all $x, y\in\mathcal{T}$, $x\succeq y$ or $y\succeq x$);
    \item $\succeq$ is transitive (i.e. for all $x, y, z \in \mathcal{T}$, if $x\succeq y$ and $y \succeq z$, then $x \succeq z$).
  \end{itemize}
  Let $\mathfrak{R}(\mathcal{T})$ be the set of all rankings on $\mathcal{T}$.
\end{definition}

In this paper, we give $x\succeq y$ the meaning that $x$ is at least as acceptable as $y$. This may be more intuitive than that of \cite{ref-amgoud2013ranking}, in which the meaning of $x\succeq y$ is just the opposite. Formally, $x \simeq y$ if and only if $x \succeq y$ and $y \succeq x$, which means $x$ and $y$ are equally acceptable. Moreover, $x \succ y$, means $x$ is strictly more acceptable than $y$, if and only if $x \succeq y$ but not $y \succeq x$.

\begin{definition}[Ranking-based Semantics]
  Let $\mathbb{G}_\mathcal{X}$ be the set of all argument systems with finite argument set $\mathcal{X}$. A ranking-based semantics is a function $\mathrm{\Gamma}: \mathbb{G}_\mathcal{X} \mapsto \mathfrak{R}(\mathcal{X})$.
\end{definition}

In other words, for a given argumentation framework $AF=\left< \mathcal{X}, \mathcal{R}\right>$, the ranking-based semantics $\mathrm{\Gamma}$ will transform $\mathcal{X}$ into a ranking $\succeq^{\textit{AF}}_{\mathrm{\Gamma}} \in \mathfrak{R}(\mathcal{X})$.

Generally, the ranking on arguments is induced by the strength values of the arguments. One of the most common approaches is categoriser function \cite{ref-besnard2001logic}, which assigns a strength value to each argument. We will discuss it in the next section.

\section{Argument ranking with categoriser function}

\subsection{Categoriser function for strength valuation}
``Categoriser'' function is originally used for ``deductive'' arguments, where an argument is structured as a pair $\left<\Phi,\phi\right>$, where $\Phi$ is a set of formulae, called \textbf{premise}, $\phi$ is a formula, called \textbf{claim}, and $\Phi$ entails $\phi$. The attack relation considered here is canonical undercut and cycles are not allowed. The notion of an ``argument tree'' captures a precise and complete representation of attackers and defenders of a given argument, root of the tree. Then, categoriser function assigns a value to a tree of arguments. This value represents the relative strength of an argument (root of the tree) given all its attackers and defenders. The categoriser function, denoted by $C$, is defined as
\begin{equation} \label{Eqn_CateFunctions}
  C(x_i) =\left\{
  \begin{array}{cl}
    1 & \mbox{~~if}~\mathcal{R}^-(x_i)= \emptyset \\
    \frac{1}{1+C(x'_1)+\cdots+C(x'_n)} & \mbox{~~if~} \mathcal{R}^-(x_i) \neq \emptyset \mbox{~with~}\mathcal{R}^-(x_i)=\{x'_1,\cdots,x'_n\}\\
  \end{array}
\right.
\end{equation}
Intuitively, the larger the number of defeaters of an argument, the lower its value. The larger the number of defenders of an argument, the larger its value.

Note that, in the work of \cite{ref-besnard2001logic}, categoriser function solely handles acyclic graphs. However, Cayrol et al. reveal that the categoriser function is an instance of their generic local valuation \cite{ref-cayrol2005graduality}, thus making it possible to cope with cycles. In this case, the strength values are the solution of non-linear equations. Specifically, let $\left< \mathcal{X}, \mathcal{R}\right>$ be an argument system with $\mathcal{X}=\{x_1,x_2,\cdots,x_n\}$, and its attack matrix be $\mathbf{D}$, and suppose the strength values of all arguments be column vector $\bm{v}$, of which the $i$-th component, denoted by $v_i$ or $\bm{v}(x_i)$, represents the strength value of $x_i$, then the strength values are the solution of the following $n$ equations:
\begin{equation}\label{Eqn_CateEquations}
  v_i = 1 / (1+\mathbf{D}_{i\ast}\cdot \bm{v}),~~~ i=1,2,\cdots,n
\end{equation}

\begin{remark}
 \eqref{Eqn_CateEquations} exactly expresses the categoriser functions in \eqref{Eqn_CateFunctions} irrespective of whether $\mathcal{R}^-(x_i)$ is empty or not, as the item $\mathbf{D}_{i\ast}\cdot\bm{v}$ exactly indicates the sum of the strength values of all attackers of $x_i$.
\end{remark}

\begin{remark}
 We merely consider all strength values as nonnegative real numbers, i.e., $v_i \geq 0$ for all $i$. Combining with \eqref{Eqn_CateEquations}, we can easily know $v_i \in [0,1]$ (actually $v_i \in (0,1]$). This means that if the solution of \eqref{Eqn_CateEquations} exists, it must be in $[0,1]^n$.
\end{remark}

\subsection{Fixed point schema for categoriser equations}
In \cite{ref-cayrol2005graduality}, the authors show a simple example to evaluate arguments in a isolated cycle with categoriser valuation by solving second degree equations. For a complex argumentation system, however, no details are available about these questions: Do these equations always exist solutions in the reals and how many real solutions exist? If the real solutions exist, how should we find them? In this subsection, we will address these questions through fixed-point techniques.


Firstly, let us transform the equations into the fixed-point form \cite{ref-burdennumerical}:
\begin{equation}\label{Eqn_CateFixedPoint}
  \bm{v} = \bm{F}(\bm{v})=[f_1(\bm{v}),f_2(\bm{v}),\cdots,f_n(\bm{v})]^T
\end{equation}
where function $\bm{F}$ maps $[0,1]^n$ into $[0,1]^n$, and the function $f_i$ from $[0,1]^n$ to $[0,1]$, called the coordinate function of $\bm{F}$, is defined by the categoriser function, i.e.,
\begin{equation}  \label{Eqn_Coordinate}
  f_i(\bm{v}) = 1 / (1+\mathbf{D}_{i\ast}\cdot \bm{v})
\end{equation}
Intuitively, $\bm{F}(\mathbf{0})= \mathbf{1}$, where the bold $\mathbf{0}$ (respectively, $\mathbf{1}$) is an appropriately dimensioned column vector of all $0$'s (respectively $1$'s).
Sometimes, we also write $f_i(\bm{v})$ as the function of $\bm{v}$ and $\mathbf{D}_{i\ast}$, i.e.,
\begin{equation}\label{Eqn_CoordinateFunction}
f_i(\bm{v})=f(\bm{v},\mathbf{D}_{i\ast})
\end{equation}
Clearly, the function $f(\bm{v},\mathbf{D}_{i\ast})$ is a non-increasing function with respect to $\bm{v}$ and $\mathbf{D}_{i\ast}$. Note that $f(\bm{v},\mathbf{0})=1$ for any $\bm{v}\in[0,1]^n$.

We convert the original problem into a fixed point problem. Then, finding the solutions of the categoriser equations is equivalent to finding the fixed-points of $\bm{F}$. In other words, a fixed-point of $\bm{F}$ is a solution of \eqref{Eqn_CateEquations}. Now, let us give the following theorem, which shows that the solution of categoriser equations always exists.

\begin{theorem}[Existence of categoriser valuation] \label{Thrm_Existence}
For any argumentation framework $AF=\left< \mathcal{X}, \mathcal{R}\right>$ with $\mathcal{X}=\{x_1,x_2,\cdots,x_n\}$, the categoriser valuation defined in \eqref{Eqn_CateEquations} has at least one solution in $[0,1]^n$.
\end{theorem}
\begin{proof}[Sketch]
  We prove the equivalence result that function $\bm{F}$ has at least one fixed point. The proof uses Brouwer's fixed point theorem \cite[Thm~2.14, pp.~24]{ref-teschl2001nonlinear} and the observation that $[0,1]^n$ is homeomorphic to a closed ball (closed, bounded, connected and without holes) and function $\bm{F}$ is continuous on it.  \qed
\end{proof}

The previous theorem is of utmost importance if we are to widely use categoriser valuation, since one would be turned away from an argumentation system that is not capable of assigning meaningful strength values (real solution) to arguments in all case.

Next we focus on the existence of a unique valuation. Assigning multiple solutions to an argumentation framework may be more interesting from a theoretical perspective, but we look forward to the kind of users of this framework to expect a unique valuation. One intuitive application of a unique categoriser valuation is that it may help removing ambiguity on argument ranking. We will show that for every argumentation system there always exists a unique categoriser valuation, and the valuation can be calculated by fixed point iteration.

\begin{theorem}[Uniqueness of categoriser valuation]  \label{Thrm_Uniqueness}
Let $AF=\left< \mathcal{X}, \mathcal{R}\right>$ be an argument system with $\mathcal{X}=\{x_1,x_2,\cdots,x_n\}$. Then, the categoriser equations defined in \eqref{Eqn_CateEquations} has a unique solution $\bm{v}^\ast \in [0,1]^n$, which is the limit of the sequence of $\{\bm{v}^{(k)}\}^\infty_{k=0}$, defined from an arbitrarily selected $\bm{v}^{(0)}$ in $[0,1]^n$ and generated by
\begin{equation}\label{Eqn_CateFixedPIteration}
  \bm{v}^{(k)} = \bm{F}(\bm{v}^{(k-1)}), \mbox{\rm ~~for each~} k \geq 1
\end{equation}
\end{theorem}
\begin{proof}
Let $\bm{u}^{(0)}=\mathbf{0}$, $\bm{u}^{(1)}=\bm{F}(\bm{u}^{(0)})=\mathbf{1}$ and $\bm{u}^{(k)}=\bm{F}(\bm{u}^{(k-1)})$ for each $k \geq 2$. Then, we can easily know that
\begin{equation}\label{Eqn_Condition1}
  \bm{u}^{(0)} \leq \bm{u}^{(2)} \leq \bm{u}^{(1)}
\end{equation}
and that there exists $0 < \varphi < 1$ such that
\begin{equation}\label{Eqn_Condition2}
  \varphi\bm{u}^{(1)}\leq\bm{u}^{(2)}
\end{equation}
Since $\bm{F}$ is non-increasing (i.e., for any $\bm{u},\bm{v}\in [0,1]^n$, if $\bm{u} \leq \bm{v}$ then $\bm{F}(\bm{u}) \geq \bm{F}(\bm{v})$), by applying $\bm{F}$ on \eqref{Eqn_Condition1} and by induction, it is easy to see that
\begin{equation}\label{Eqn_SequenceU}
  \mathbf{0} = \bm{u}^{(0)} \leq \bm{u}^{(2)} \leq \cdots \leq \bm{u}^{(2k)} \leq \cdots \leq \bm{u}^{(2k+1)} \leq \cdots \leq \bm{u}^{(3)} \leq \bm{u}^{(1)} = \mathbf{1}
\end{equation}
On the other hand, from \eqref{Eqn_Condition2} and \eqref{Eqn_SequenceU}, we find $\varphi\bm{u}^{(2k-1)}\leq\bm{u}^{(2k)}$ for each $k\geq 1$. Letting $\pi_k = \sup\{\pi: \pi\bm{u}^{(2k-1)} \leq \bm{u}^{(2k)}\}$, then $\pi_k\bm{u}^{(2k-1)} \leq \bm{u}^{(2k)}$ and $0<\varphi\leq \pi_1 \leq \cdots \leq \pi_k \leq \cdots \leq 1$. In the following, we prove that $\lim_{k\rightarrow\infty} \pi_k = 1$.

Note that $f_i(\pi\bm{u})=\frac{1}{\pi+f_i(\bm{u})(1-\pi)}f_i(\bm{u})$ for all $i\in\{1,2,\cdots,n\}$, then there exists $0<\alpha<1$ and a continuous function $\psi(\pi)=\frac{1}{\pi+\alpha(1-\pi)}$ such that
\begin{equation}\label{Eqn_Condition3}
  \bm{F}(\pi\bm{u})\leq\psi(\pi)\bm{F}(\bm{u}), ~~\forall \pi\in[\varphi,1), \bm{u}\in[\varphi,1]^n
\end{equation}
Then, by \eqref{Eqn_SequenceU}, \eqref{Eqn_Condition3} and the non-increasing property of $\bm{F}$, we have
\begin{equation}\label{Eqn_PI_Limit}
 \bm{u}^{(2k+1)} = \bm{F}(\bm{u}^{(2k)}) \leq \bm{F}(\pi_k\bm{u}^{(2k-1)}) \leq \psi(\pi_k)\bm{u}^{(2k)} \leq \psi(\pi_k)\bm{u}^{(2k+2)}
\end{equation}
which implies that $\pi_{k+1} \geq \frac{1}{\psi(\pi_k)} = \pi_k+\alpha(1-\pi_k)$. So,
\begin{equation}\label{Eqn_Error}
  1-\pi_{k+1} \leq (1-\alpha)(1-\pi_k) \leq \cdots \leq (1-\alpha)^k(1-\pi_1) \leq (1-\alpha)^k(1-\varphi)
\end{equation}
As $0<\alpha<1$, thus by \eqref{Eqn_Error} we have
\begin{equation}\label{Eqn_Limits1}
  \lim_{k\rightarrow\infty}(1-\pi_{k+1})=0 ~~~~\Rightarrow~~~~ \lim_{k\rightarrow\infty} \pi_k = 1
\end{equation}
Therefore, by \eqref{Eqn_SequenceU} we get, for any integer $p\geq 1$
\begin{equation}\label{Eqn_Error2}
  \mathbf{0} \leq \bm{u}^{(2k+2p)}-\bm{u}^{(2k)} \leq \bm{u}^{(2k+1)}-\bm{u}^{(2k)} \leq (1-\pi_k)\bm{u}^{(2k+1)} \leq (1-\pi_k)\bm{u}^{(1)}
\end{equation}
Since $[0,1]^n$ is normal, both $\{\bm{u}^{(2k+1)}\}^\infty_{k=0}$ and $\{\bm{u}^{(2k)}\}^\infty_{k=1}$ are convergence sequences. By \eqref{Eqn_Limits1} and \eqref{Eqn_Error2}, thus, there exists $\bm{u}^\ast\in[0,1]^n$ such that
\begin{equation}\label{Eqn_Limits2}
  \lim_{k\rightarrow\infty}\bm{u}^{(2k+1)}=\lim_{k\rightarrow\infty}\bm{u}^{(2k)}=\bm{u}^\ast
\end{equation}
Hence $\bm{u}^{(2k)}\leq\bm{u}^\ast\leq\bm{u}^{(2k-1)}$ and $\bm{u}^{(2k)}\leq\bm{F}(\bm{u}^\ast)\leq\bm{u}^{(2k+1)}$. Letting $k\rightarrow\infty$ and combining with \eqref{Eqn_Limits2}, it follows $\bm{F}(\bm{u}^\ast)=\bm{u}^\ast$, i.e., $\bm{u}^\ast$ is a fixed point of $\bm{F}$.

Now, for any $\bm{v}^{(0)}\in[0,1]^n$ and for any $k\geq 1$, by induction, we have $\bm{u}^{(2k)}\leq\bm{v}^{(2k)}\leq\bm{u}^{(2k-1)}$ and $\bm{u}^{(2k)}\leq\bm{v}^{(2k+1)}\leq\bm{u}^{(2k+1)}$. Then $\bm{v}^{(k)}\rightarrow \bm{v}^\ast = \bm{u}^\ast$ as $k\rightarrow\infty$. In particular, let $\bm{v}^{(0)}=\bm{w}^\ast$, where $\bm{w}^\ast$ is any fixed point of $\bm{F}$ in $[0,1]^n$, then $\bm{v}^{(k)}=\bm{w}^\ast$ for all $k\geq 1$, and we get $\bm{w}^\ast=\bm{u}^\ast$. So, $\bm{F}$ has a unique fixed point in $[0,1]^n$.
\qed
\end{proof}

\begin{algorithm}[tb] \label{Alg_IterativeValuation}
\KwIn{$\mathbf{D}_{n\times n}$: attack matrix; $\epsilon$: prescribed tolerance;}
\KwOut{$\bm{v}^{(k)}$: the approximate solution of the categoriser equations}
\Begin{
$k \longleftarrow 0$; $\bm{v}^{(0)} \longleftarrow \mathbf{1}$\;

\Repeat{$\|\bm{v}^{(k)}-\bm{v}^{(k-1)}\| \leqslant \epsilon$}{
$k \longleftarrow k+1$\;
$v^{(k)}_{i} = f(\bm{v}^{(k-1)}, \mathbf{D}_{i\ast}) \mbox{~~for each~} i\in\{1,2,\cdots,n\}$\;
}
\Return $\bm{v}^{(k)}$\;
}
\caption{Fixed-point iteration for categoriser valuation}
\end{algorithm}

The proof of this theorem mainly refers to \cite[Lmm~2.1]{ref-li2005positive}. An approximate calculation of the unique categoriser valuation $\bm{v}^\ast$ is done by using Algorithm~\ref{Alg_IterativeValuation}. In this paper, we set the initial strength values $\bm{v}^{(0)}=\mathbf{1}$ since we assume that each argument is not attacked at the beginning and has the maximum strength value $1$. The iteration terminates when the change of the sequence $\{\bm{v}^{(k)}\}^\infty_{k=0}$ is under a given tolerance $\epsilon$. As the proof of uniqueness suggests, the estimation of convergence rate of this algorithm is
\begin{equation}\label{Eqn_Convergence}
  \|\bm{v}^{(2k)}-\bm{v}^\ast\|\leq \|\bm{v}^{(2k)}-\bm{u}^{(2k)}\|+\|\bm{v}^\ast-\bm{u}^{(2k)}\| \leq 2\|\bm{u}^{(2k+1)}-\bm{u}^{(2k)}\|
\end{equation}
By \eqref{Eqn_Error} and \eqref{Eqn_Error2}, we have $\|\bm{v}^{(2k)}-\bm{v}^\ast\|\leq 2(1-\alpha)^{k-1}(1-\varphi)\|\bm{u}^{(1)}\|$. Similar argument gives that $\|\bm{v}^{(2k+1)}-\bm{v}^\ast\|\leq 2(1-\alpha)^{k-1}(1-\varphi)\|\bm{u}^{(1)}\|$.

\begin{remark}
In Algorithm~\ref{Alg_IterativeValuation}, we can see that at each iterative step the strength value of any argument $x_i$ is simultaneously recomputed in the light of its direct attackers (represented by $\mathbf{D}_{i\ast}$) and the strength values in the previous step (i.e., $\bm{v}^{(k-1)}$). This exactly embodies the idea of ``local approach'' (i.e., the value of an argument only depends on the values of its direct attackers) in \cite{ref-cayrol2005graduality}.
\end{remark}

\subsection{Categoriser-based ranking semantics}
Now, we have shown that for the categoriser equations there always exists a unique solution for any argumentation framework. The solution assigns a numerical value to each argument, which can be interpreted as the strength of the argument. The greater the strength value, the more acceptable the argument. Thus, we can induce a ranking on arguments from the unique solution.
\begin{definition}[Categoriser-based ranking semantics] \label{Def_CateRankingSemantics}
Let $AF=\left< \mathcal{X}, \mathcal{R}\right>$ be an argumentation framework, and $\bm{v}^{\ast}$ be the unique solution of \eqref{Eqn_CateEquations}. The categoriser-based ranking semantics is a ranking-based semantic and transforms $AF$ into the ranking $\succeq$ such that $\forall x_i, x_j\in \mathcal{X}$, $x_i \succeq x_j$ if and only if $\bm{v}^\ast(x_i) \geq \bm{v}^\ast(x_j)$.
\end{definition}

Obviously, the categoriser-based ranking semantics satisfies that for any $x\in\mathcal{X}$, $\bm{v}^\ast(x_i)=1$ if $\mathcal{R}^-(x_i)=\emptyset$; else $\bm{v}^\ast(x_i)<1$, i.e., non-attacked arguments are more acceptable than attacked ones. Non-attacked arguments are supported by extension-based semantics. They are part of any extension under complete, preferred, stable and grounded semantics. Therefore, it is naturel to believe that an argument which has no attackers is ranked higher than another argument which has attackers.

In addition, we give other properties of the categoriser-based ranking semantics:
\begin{proposition}  \label{Prop_SameDirectAttackers}
  Let $x_i,x_j\in\mathcal{X}$. The categoriser-based ranking semantics satisfies:
  \begin{enumerate}[{\rm [P1]:~}]
  \item If $\mathcal{R}^-(x_i)=\mathcal{R}^-(x_j)$, then $x_i \simeq x_j$.
  \item If $\mathcal{R}^-(x_i) \subseteq \mathcal{R}^-(x_j)$, then $x_i \succeq x_j$.
\end{enumerate}
\end{proposition}
\begin{proof}
By \eqref{Eqn_CateFixedPIteration}, the categoriser strength of any argument $x_i$ can be written as
\begin{equation} \label{Eqn_CategoriserStrengthOfXi}
  \bm{v}^\ast(x_i) = \lim_{k \rightarrow \infty} \bm{v}^{(k)}(x_i) = \lim_{k \rightarrow \infty} f_i(\bm{v}^{(k-1)}) = \lim_{k \rightarrow \infty} f(\bm{v}^{(k-1)}, \mathbf{D}_{i\ast})
\end{equation}
For {\rm [P1]}, $\mathcal{R}^-(x_i)=\mathcal{R}^-(x_j)$ implies that $\mathbf{D}_{i\ast}=\mathbf{D}_{j\ast}$, which implies $f(\bm{v}^{(k-1)}, \mathbf{D}_{i\ast})=f(\bm{v}^{(k-1)}, \mathbf{D}_{j\ast})$. By \eqref{Eqn_CategoriserStrengthOfXi}, we have $\bm{v}^\ast(x_i)=\bm{v}^\ast(x_j)$, i.e., $x_i\simeq x_j$. For {\rm [P2]}, $\mathcal{R}^-(x_i)\subseteq\mathcal{R}^-(x_j)$ means that $\mathbf{D}_{i\ast} \leq \mathbf{D}_{j\ast}$. Since $f(\bm{v},\mathbf{D}_{i\ast})$ is a non-increasing function of $\mathbf{D}_{i\ast}$, we have $f(\bm{v}^{(k-1)}, \mathbf{D}_{i\ast}) \geq f(\bm{v}^{(k-1)}, \mathbf{D}_{j\ast})$. Thus, $\bm{v}^\ast(x_i)\geq\bm{v}^\ast(x_j)$, i.e., $x_i\succeq x_j$.
  \qed
\end{proof}
This proposition states that two arguments with the same direct attackers have the same ranking, and an argument, whose direct attackers pertain to the set of direct attackers of another argument, is at least as more acceptable than the argument.

Let us show an example of how the semantics works:
\begin{example} \label{Exp_ValuationSeq}
Consider again the argument system in Fig.~\ref{Fig_ExampArg}. Let $\epsilon = 10^{-3}$ and $\bm{v}^{(0)} = \mathbf{e}$. Then, the valuation sequence $\{\bm{v}^{(k)}\}^\infty_{k=0}$, calculated by Algorithm~\ref{Alg_IterativeValuation}, is shown in Fig.~\ref{Fig_IterValuation}.

When $k=0$, all arguments have the maximum strength value $1$ as we presuppose each argument is not attacked at the beginning.

When $k=1$, then the strength value of each argument merely depends on the number of its direct attackers since the strength values of all argument from the previous step are $1$. Thus, $x_3$ has the maximum strength value $1$ since it has no attacker, followed by $x_1$ with one attacker, and followed by $x_4$ and $x_5$ with two attackers, and followed by $x_2$ with three. From another perspective, since $\mathcal{R}^-(x_3) \subset\mathcal{R}^-(x_1) \subset \mathcal{R}^-(x_5) \subset \mathcal{R}^-(x_2)$, then by Proposition~\ref{Prop_SameDirectAttackers} we have $x_3 \succ x_1 \succ x_5 \succ x_2$.

When $k=2$, after a new round of calculation, the strength value of each argument is recomputed. But, since $\mathcal{R}^-(x_3) \subset\mathcal{R}^-(x_1) \subset \mathcal{R}^-(x_5) \subset \mathcal{R}^-(x_2)$ always holds, the ranking among them will not be changed. Note that the ranking on $x_2$ and $x_4$ is altered as the sum of the strength values of the attackers of $x_2$ is greater than that of $x_4$.

......

After finitely many iterations, the valuation sequence gradually tends to be stable and converge to an approximative solution $\bm{v}^\ast=[0.72,0.43,1.00,0.40,0.51]^T$ within a tolerable range. Actually, the valuation sequence reflects how argument strength values change with iterations. Note that $x_1$ has a maximum strength value $1$, since it is not attacked, and all other arguments have the strength values less than $1$ since they are attacked by at least one argument. With the solution $\bm{v}^\ast$, the categoriser-based ranking semantics gives: $x_3\succ x_1 \succ x_5 \succ x_2 \succ x_4$.
\begin{figure}[tb]
\centering
  \includegraphics[width=0.75\textwidth]{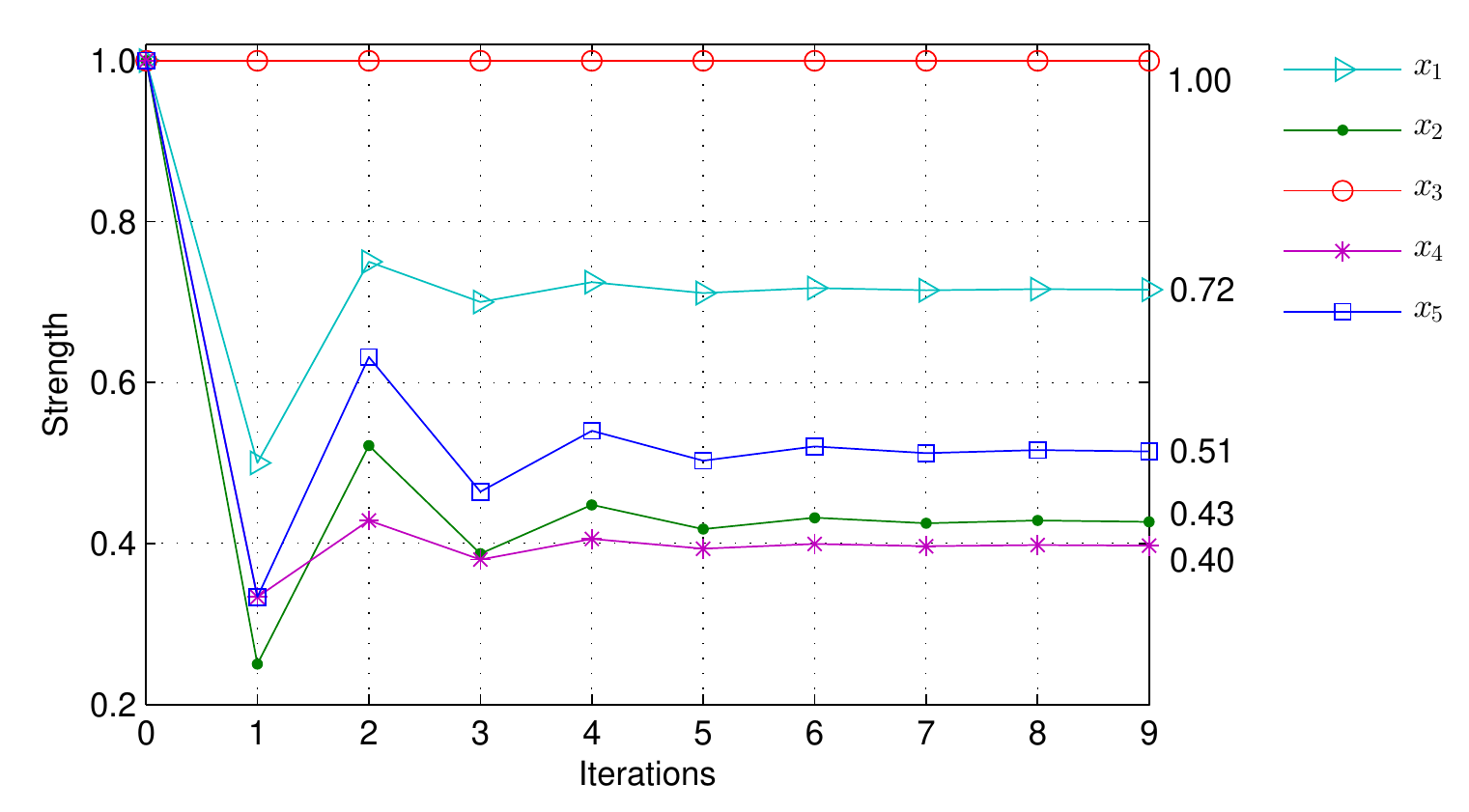}
\caption{Categoriser valuation sequence of Example~\ref{Exp_ArgNetwork}}
\label{Fig_IterValuation}
\vspace{-0.4cm}
\end{figure}
\end{example}

\section{Relating with ranking axioms}
In \cite{ref-amgoud2013ranking}, the authors set up a set of axiom (postulates) that ranking-based semantics should satisfy. In this section, we will formally show that the categoriser-based ranking semantics meets some of these postulates.

The first axiom is that a ranking on a set of arguments does not rely on their identity but only on the attack relations among them. In other words, if two argumentation system are isomorphic then they should have the same ranking semantics. The isomorphisms between argumentation frameworks $\textit{AF}_1=\left< \mathcal{X}_1, \mathcal{R}_1\right>$ and $\textit{AF}_2=\left< \mathcal{X}_2, \mathcal{R}_2\right>$ is a bijective function $\tau$: $\mathcal{X}_1 \mapsto \mathcal{X}_2$ such that for all $x,y\in \mathcal{X}_1$, $x\mathcal{R}_1 y$ if and only if $\tau(x)\mathcal{R}_2\, \tau(y)$. Now we define the first axiom, called \textit{abstraction}, as follows:
\begin{axiom}[Abstraction {\sf (Ab)}]
  A ranking-based semantics $\mathrm{\Gamma}$ satisfies {\rm abstraction} iff for any two argumentation framework $\textit{AF}_1=\left< \mathcal{X}_1, \mathcal{R}_1\right>$ and $\textit{AF}_2=\left< \mathcal{X}_2, \mathcal{R}_2\right>$, for any isomorphism $\tau$ from $\textit{AF}_1$ to $\textit{AF}_2$, we have $\forall x,y\in \mathcal{X}_1$, $x \succeq^{\textit{AF}_1}_{\mathrm{\Gamma}} y$ iff $\tau(x) \succeq^{\textit{AF}_2}_{\mathrm{\Gamma}} \tau(y)$.
\end{axiom}

The second axiom states the question that whether an argument $x$ is at least as acceptable as an argument $y$ should be independent of any argument $z$ that is not connected to $x$ or $y$, i.e., there is no path from $x$ or $y$ to $z$ (neglecting the direction of the edges). Let $\mathcal{C}(\textit{AF})$ be the set of weakly connected components of $\textit{AF}$. Each weakly connected component of $\textit{AF}$ is a maximal subgraph of $\textit{AF}$ in which any two arguments are mutually connected by a path (neglecting the direction of the edges).
\begin{axiom}[Independence {\sf (In)}]
  A ranked-based semantics $\mathrm{\Gamma}$ satisfies {\rm independence} iff for any $\textit{AF}$ and for any $\textit{AF}_c=\left< \mathcal{X}_c, \mathcal{R}_c\right> \in 2^{\mathcal{C}(\textit{AF})}$, $\forall x,y\in\mathcal{X}_c$, $x \succeq^{\textit{AF}}_{\mathrm{\Gamma}} y$ iff $x \succeq^{\textit{AF}_c}_{\mathrm{\Gamma}} y$.
\end{axiom}

The third axiom, called \textit{void precedence}, encodes the idea that non-attacked arguments are more acceptable than attacked ones.
\begin{axiom}[Void Precedence {\sf (VP)}]
   A ranked-based semantics $\mathrm{\Gamma}$ meets {\rm void precedence} iff for any $\textit{AF}=\left< \mathcal{X}, \mathcal{R}\right>$, $\forall x,y\in\mathcal{X}$, if $\mathcal{R}^-(x)=\emptyset$ and $\mathcal{R}^-(y)\neq\emptyset$ then $x \succeq^{\textit{AF}}_{\mathrm{\Gamma}} y$.
\end{axiom}

The fourth axiom states that having attacked attackers is more acceptable than non-attacked attackers, i.e., being defended is better than not.
\begin{axiom}[Defense precedence {\sf (DP)}]
A ranked-based semantics $\mathrm{\Gamma}$ satisfies {\rm defense pre-cedence} iff for every $\textit{AF}=\left< \mathcal{X}, \mathcal{R}\right>$, $\forall x,y\in\mathcal{X}$, if $|\mathcal{R}^-(x)|=|\mathcal{R}^-(y)|$, $\mathcal{D}(x)\neq \emptyset$ and $\mathcal{D}(y)=\emptyset$ then $x \succeq^{\textit{AF}}_{\mathrm{\Gamma}} y$.
\end{axiom}

The next axiom says that an argument $x$ should be at least as acceptable as argument $y$, when the direct attackers of $y$ are at least as numerous and well-ranked as those of $x$. This involves the concept of \textit{group comparison}: Let $\succeq_{\mathrm{\Gamma}}$ be a ranking on a set of arguments $\mathcal{X}$. For any $S_1,S_2\subseteq \mathcal{X}$, $S_1 \succeq_{\mathrm{\Gamma}} S_2$ iff there exists an injective mapping $\delta$ from $S_2$ to $S_1$ such that $\forall x\in S_2$, $\delta(x)\succeq_{\mathrm{\Gamma}}x$. Moreover, $S_1 \succ_{\mathrm{\Gamma}} S_2$ is a \textit{strict group comparison} iff (1) $S_1 \succeq_{\mathrm{\Gamma}} S_2$; (2) $|S_2|<|S_1|$ or $\exists x \in S_2$, $\delta(x)\succ_{\mathrm{\Gamma}} x$.
\begin{axiom}[Counter-Transitivity {\sf (CT)}]
  A ranked-based semantics $\mathrm{\Gamma}$ satisfies {\rm counter-transitivity} iff for every $\textit{AF}=\left< \mathcal{X}, \mathcal{R}\right>$, $\forall x,y\in\mathcal{X}$, if $\mathcal{R}^-(y) \succeq^{\textit{AF}}_{\mathrm{\Gamma}} \mathcal{R}^-(x)$ then $x \succeq^{\textit{AF}}_{\mathrm{\Gamma}} y$.
\end{axiom}
\begin{axiom}[Strict Counter-Transitivity {\sf (SCT)}]
  A ranked-based semantics $\mathrm{\Gamma}$ satisfies {\rm strict} {\sf (CT)} iff for any $\textit{AF}=\left< \mathcal{X}, \mathcal{R}\right>$, $\forall x,y\in\mathcal{X}$, if $\mathcal{R}^-(y) \succ^{\textit{AF}}_{\mathrm{\Gamma}} \mathcal{R}^-(x)$ then $x \succ^{\textit{AF}}_{\mathrm{\Gamma}} y$.
\end{axiom}

The following two axioms represent two opinions: give precedence to cardinality over quality (i.e. two weakened attackers is worse for the target than one strong attacker), or vice versa. In some situations, both choices are reasonable.
\begin{axiom}[Cardinality Precedence {\sf (CP)}]
  A ranked-based semantics $\mathrm{\Gamma}$ satisfies {\rm cardinality precedence} iff for arbitrary argumentation framework $\textit{AF}=\left< \mathcal{X}, \mathcal{R}\right>$, $\forall x,y\in\mathcal{X}$, if $|\mathcal{R}^-(x)| < |\mathcal{R}^-(y)|$ then $x \succ^{\textit{AF}}_{\mathrm{\Gamma}} y$.
\end{axiom}
\begin{axiom}[Quality Precedence {\sf (QP)}]
    A ranked-based semantics $\mathrm{\Gamma}$ satisfies {\rm quality precedence} iff for arbitrary argumentation framework $\textit{AF}=\left< \mathcal{X}, \mathcal{R}\right>$, $\forall x,y\in\mathcal{X}$, if $\exists y' \in \mathcal{R}^-(y)$ such that $\forall x'\in\mathcal{R}^-(x)$, $y' \succ^{\textit{AF}}_{\mathrm{\Gamma}} x'$, then $x \succ^{\textit{AF}}_{\mathrm{\Gamma}} y$.
\end{axiom}

The last axiom focuses on the way arguments are defended. The main idea is that an argument which is defended against more attackers is more acceptable than an argument which is defended against a smaller number of attacks. There are two types of defense: simple and distributed. The defense of an argument $x$ is \emph{simple} iff each defender of $x$ attacks exactly one attacker of $x$, formally, $\forall y \in \mathcal{D}(x)$ such that $|\mathcal{R}^+(y)\cap\mathcal{R}^-(x)|=1$. The defense of an argument $x$ is \emph{distributed} iff every attacker of $x$ is attacked by at least one argument, i.e., $\forall y \in \mathcal{R}^-(x)$ such that $|\mathcal{R}^-(y)|\geq 1$.
\begin{axiom}[Distributed-Defense Precedence {\sf (DDP)}]
A ranked-based semantics $\mathrm{\Gamma}$ satisfies \emph{distributed-defense precedence} iff for any $\textit{AF}=\left< \mathcal{X}, \mathcal{R}\right>$, $\forall x,y\in\mathcal{X}$ such that $|\mathcal{R}^-(x)|=|\mathcal{R}^-(y)|$ and $|\mathcal{D}(x)|=|\mathcal{D}(y)|$, if the defense of $x$ is simple and distributed and the defense of $y$ is simple but not distributed then $x \succ^{\textit{AF}}_{\mathrm{\Gamma}} y$.
\end{axiom}

In addition, \cite{ref-amgoud2013ranking} provides some relationships between these axioms: if a ranking-based semantics $\mathrm{\Gamma}$ satisfies {\sf (SCT)} then it satisfies {\sf (VP)}; if $\mathrm{\Gamma}$ satisfies both {\sf (CT)} and {\sf (SCT)}, then it satisfies {\sf (DP)}; $\mathrm{\Gamma}$ can not satisfy both {\sf (CP)} and {\sf (QP)}. Now, we show which axioms are or are not compatible with the categoriser-based ranking semantics.
\begin{theorem}
  The categoriser-based ranking semantics satisfies {\sf (Ab)}, {\sf (In)}, {\sf (VP)}, {\sf (DP)}, {\sf (CT)} and {\sf (SCT)}, and does not satisfy {\sf (CP)}, {\sf (QP)} and {\sf (DDP)}.
\end{theorem}
From the definition of the categoriser function, it can be easily seen that categoriser-based ranking semantics satisfies {\sf (Ab)} and {\sf (In)}. To some extent, Proposition~\ref{Prop_SameDirectAttackers} is a special case of {\sf (CT)}. In particular, when $\mathcal{R}^-(x_i)\subset\mathcal{R}^-(x_j)$ then the semantics gives $x_i\succ x_j$, which is a special case of {\sf (SCT)}. The {\sf (VP)} and {\sf (DP)} can be implied from {\sf (CT)} and {\sf (SCT)}. Now, let us give a counter example to show that the semantics does not satisfy {\sf (CP)} and {\sf (QP)}:
\begin{example}
Consider the argument system in Fig.~\ref{Fig_CounterExample}, in which $\mathcal{R}^-(x)=\{x_1,x_2,x_3\}$ and $\mathcal{R}^-(y)=\{y_1,y_2,y_3,y_4\}$. Clearly, $|\mathcal{R}^-(x)|<|\mathcal{R}^-(y)|$. However, the categoriser-based ranking semantics gives $y \succ x$ (since $\bm{v}^\ast(x)=0.40$ and $\bm{v}^\ast(y)=0.43$), which conflicts with {\sf (CP)}. Note that $y_1 \succ x_i$ for all $i\in\{1,2,3\}$ (since $y_1=0.60$ and $x_i=0.50$). From {\sf (QP)}, $x\succ y$ should hold, but it is not true for the semantics.

\begin{figure}[htb]
\vspace{-0.4cm}
\centering
 \includegraphics[width=0.95\textwidth]{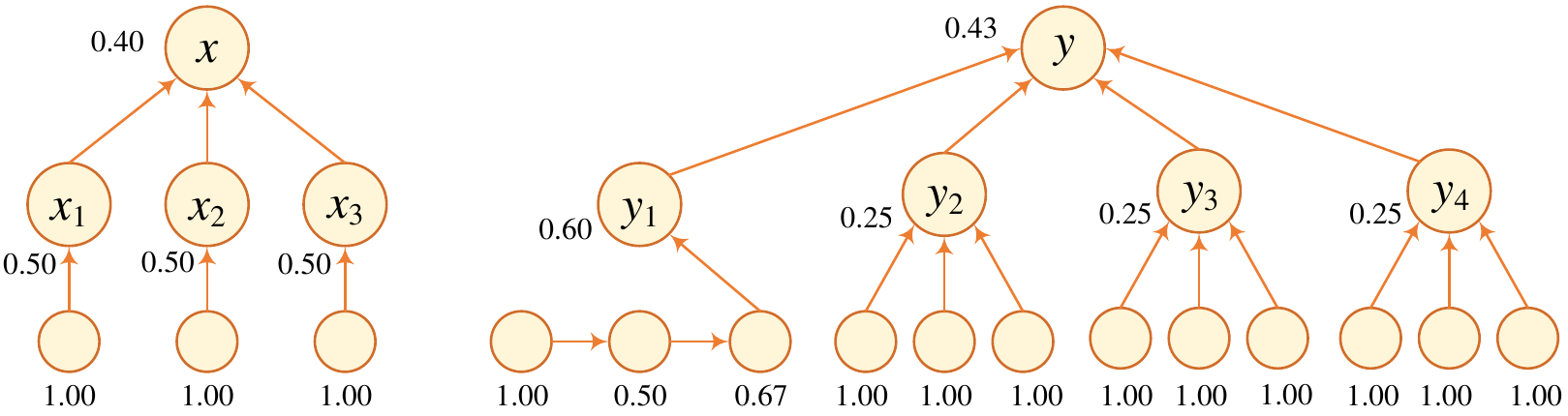}
\caption{A counter-example of axiom {\sf (CP)} and {\sf (QP)}}  \label{Fig_CounterExample}
\vspace{-0.4cm}
\end{figure}
\end{example}

The main reason of the counter situation in the above example is that these two axioms represent two extreme: one treats all attackers equally, and one merely focuses on some attacker (with highest rank with respect to the set of attackers of the argument) of an argument. In categoriser valuation, however, the value of an argument (represented by $f(\bm{v},\mathbf{D}_{i\ast})$) depends on both the number and quality (i.e., the strength values) of its attackers, the attackers of its attackers, etc.

Another reason that {\sf (CP)} is not satisfied by the categoriser valuation is that {\sf (CP)} concentrates too much on quite local topological aspects of an argumentation framework, but ignores the global topology \cite{ref-thimm2013stratified}. However, the categoriser valuation is a global approach since the strength value of an argument depends on the strength values of its attackers, which is a recursive definition. For the same reason, the categoriser-based ranking semantics does not satisfy {\sf (DDP)}.

\section{Conclusion}
In this paper, we firstly investigated the existence and uniqueness of the categoriser strength valuation via fixed-point technique. On this basis, we then defined a new ranking-based semantics, called categoriser-based ranking semantics, for abstract argumentation framework. We analyzed some general properties of the semantics, and prove that it satisfies some of the postulates that a ranking-based semantics should satisfy. Our ongoing work is about a deeper analysis of the approach and its relationships to other approaches.


\bibliographystyle{splncs}
\bibliography{sigproc}
\end{document}